\documentclass[10pt,twocolumn,letterpaper]{article}

\usepackage[pagenumbers]{cvpr} 

\usepackage{graphicx}
\usepackage{amsmath}
\usepackage{amssymb}
\usepackage{booktabs}
\usepackage{breqn}
\usepackage{tablefootnote}

\usepackage[pagebackref,breaklinks,colorlinks]{hyperref}

\usepackage[capitalize]{cleveref}
\crefname{section}{Sec.}{Secs.}
\Crefname{section}{Section}{Sections}
\Crefname{table}{Table}{Tables}
\crefname{table}{Tab.}{Tabs.}

\def\cvprPaperID{11} 
\def\confName{CVPR}
\def\confYear{2022}

\begin{document}

\title{Depth Estimation with Simplified Transformer}

\author{John Yang, Le An, Anurag Dixit, Jinkyu Koo, Su Inn Park\\
NVIDIA\\
{\tt\small \{johnyang, lean, anuragd, jinkyuk, joshp\}@nvidia.com}
}
\maketitle

\begin{abstract}
Transformer and its variants have shown state-of-the-art results in many vision tasks recently, ranging from image classification to dense prediction. Despite of their success, limited work has been reported on improving the model efficiency for deployment in latency-critical applications, such as autonomous driving and robotic navigation. In this paper, we aim at improving upon the existing transformers in vision, and propose a method for self-supervised monocular Depth Estimation with Simplified Transformer (DEST), which is efficient and particularly suitable for deployment on GPU-based platforms. Through strategic design choices, our model leads to significant reduction in model size, complexity, as well as inference latency, while achieving superior accuracy as compared to state-of-the-art. We also show that our design generalize well to other dense prediction task without bells and whistles.
\end{abstract}

\section{Introduction}
\label{sec:intro}

Accurate depth estimation is an essential capability for geometric perception within a scene. Estimated depth provides rich visual cues for general perception, navigation, planning, and reasoning against occlusions for applications such as robotics \cite{eigen2014depth} and advanced driver-assistance systems (ADAS) \cite{ali2020real}. Recently, deep learning based methods \cite{eigen2014depth, liu2015learning} have shown that depth can be learned from a single image by using convolutional neural networks (CNN). However, direct supervision requires large amount of ground-truth depth maps, which are expensive to obtain in reality. On the other hand, self-supervised, or sometimes referred to as unsupervised methods can take advantage of geometrical constraints on image sequences as the sole source of supervision. For example, previous work~\cite{mahjourian2018unsupervised, packnet, zhao2020towards} showed that CNN-based depth and ego-motion networks can be solely trained on monocular video sequences without using ground-truth depth or stereo image pairs.

\begin{figure*}
    \centering
    \includegraphics[width=0.9\linewidth]{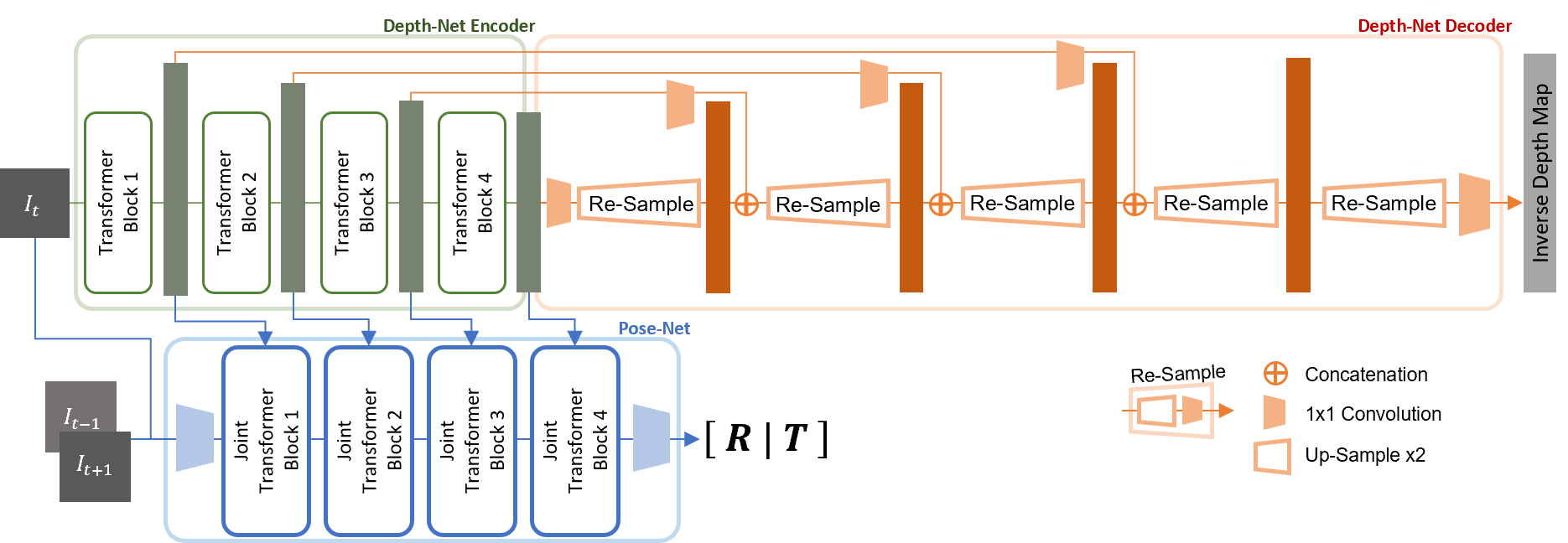}
    \vspace{-2mm}
    \caption{The proposed framework for self-supervised depth estimation with simplified transformer (DEST). Both Depth-Net and Pose-Net are trained together. For inference, only Depth-Net is needed.}
    \vskip -0.15in
    \label{fig:dest}
\end{figure*}

The key to the self-supervised learning methods is to build a task consistency for training separated CNN networks, where predictions from depth network and pose network are jointly constrained by image reconstruction error~\cite{roy2016monocular, zhou2017unsupervised, godard2017unsupervised, packnet, klodt2018supervising, bian2019unsupervised, zhao2020towards}. While this structure of paired depth-pose networks has been largely adopted, they are mainly built with CNNs that have evolved towards more complex architectures that are computationally demanding. 

On the other hand, inspired by the seminal work on transformer~\cite{vaswani2017attentionisallyouneed}, vision transformers have emerged in many applications~\cite{vit, deit, liu2021swin}, benefiting from the attention mechanism and simpler network structure. In order to exploit the capacity of vision transformers and self-supervised learning systems~\cite{caron2021emerging_dino}, in this paper we aim to improve the performance of self-supervised monocular depth estimation in terms of both accuracy and latency for deployment, through improved design choices. To this end, our main contributions are the following:
\begin{itemize}
    \item Simplified transformer with strategical design choices that are hardware friendly, yielding networks that only consist of very basic operations and operate efficiently
    \item Transformer-based Depth-Net and Pose-Net with joint attention mechanism for more effective learning
\end{itemize}

We show that on public benchmark dataset our model is over 85\% smaller in model size and complexity, while being significantly faster in terms of latency and more accurate, as compared to previous state-of-the-art. We also show that this architecture can generalize well to other dense prediction tasks such as semantic segmentation.

\section{Related Works}
\noindent \textbf{Monocular Depth Estimation.} With the emergence of learning-based methods, it has been shown that depth can be learned directly from a single image in a supervised manner~\cite{eigen2014depth, liu2015learning}. To date, self-supervised methods, which take advantage of abundant unlabeled data for training, have drawn great attention. Godard \etal~\cite{godard2017unsupervised} proposed a novel training objective that enforces consistency between disparity produced by left and right images, and this method was further improved in~\cite{monodepth2} with new choices of loss functions and training strategies. In ~\cite{packnet}, symmetrical packing and unpacking blocks are proposed to generate detailed representation using 3D convolutions along with velocity information for better scale-awareness. 

\noindent \textbf{Vision Transformers.} The pioneer work of Vision Transformer (ViT)~\cite{vit} showed that a pure transformer with sequence of image patches as input works as well as CNNs. To alleviate the need of large amount of pre-training data, Data-efficient image Transformer (DeiT)~\cite{deit} was proposed with a novel distillation strategy. To address image scale and resolution variation, Swin Transformer, a hierarchical transformer with shifted windows, was invented~\cite{liu2021swin}. Pyramid Vision Transformer (PVT)~\cite{pvt} utilizes a pyramid structure that allows fine-grained inputs for dense prediction tasks in conjunction with progressive shrinking to reduce computations. SegFormer~\cite{xie2021segformer} leverages multi-scale features for accurate pixel-level prediction, while positional encoding and complexity in decoder are dropped, resulting in performance boost in both accuracy and latency. Recently, Li \etal proposed STereo TRansformer (STTR) which employs transformer for sequence-to-sequence correspondence modeling for stereo depth estimation~\cite{li2021revisiting}. ViT backbone is employed as a encoder along with a convolutional decoder for supervised depth estimation in~\cite{Ranftl_2021_ICCV}.

\noindent \textbf{Efficient Transformers.} Some recent efforts focus on reducing the complexity in transformers mainly from a theoretical perspective~\cite{wang2020linformer, choromanski2021rethinking}. For vision tasks, Li \etal~\cite{li2021efficient} introduced a multi-stage efficient transformer (EsViT) in conjunction with sparse self-attention. Jia \etal~\cite{jia2021efficient} proposed an efficient fine-grained manifold distillation approach that allows student network to achieve better performance when learning from the more complex teacher transformer. However, in reality the projected speedup may not be realized due to hardware limitations that inherently determine software capability. Different from the scope of the aforementioned approaches, we introduce a simplified design of transformer architecture, which is particular friendly to GPU hardware in order to achieve efficient inference. 

\section{Method}
The design of Depth-Net and Pose-Net is shown in Fig.~\ref{fig:dest}. The encoder of Depth-Net and the Pose-Net consist of four transformer blocks. In the training process, the predicted depth map and pose are used for view synthesis, and the photometric loss is used as training objective~\cite{zhou2017unsupervised, packnet}. Only Depth-Net is needed for inference.

\begin{figure}
    \centering
    \includegraphics[width=0.82\linewidth]{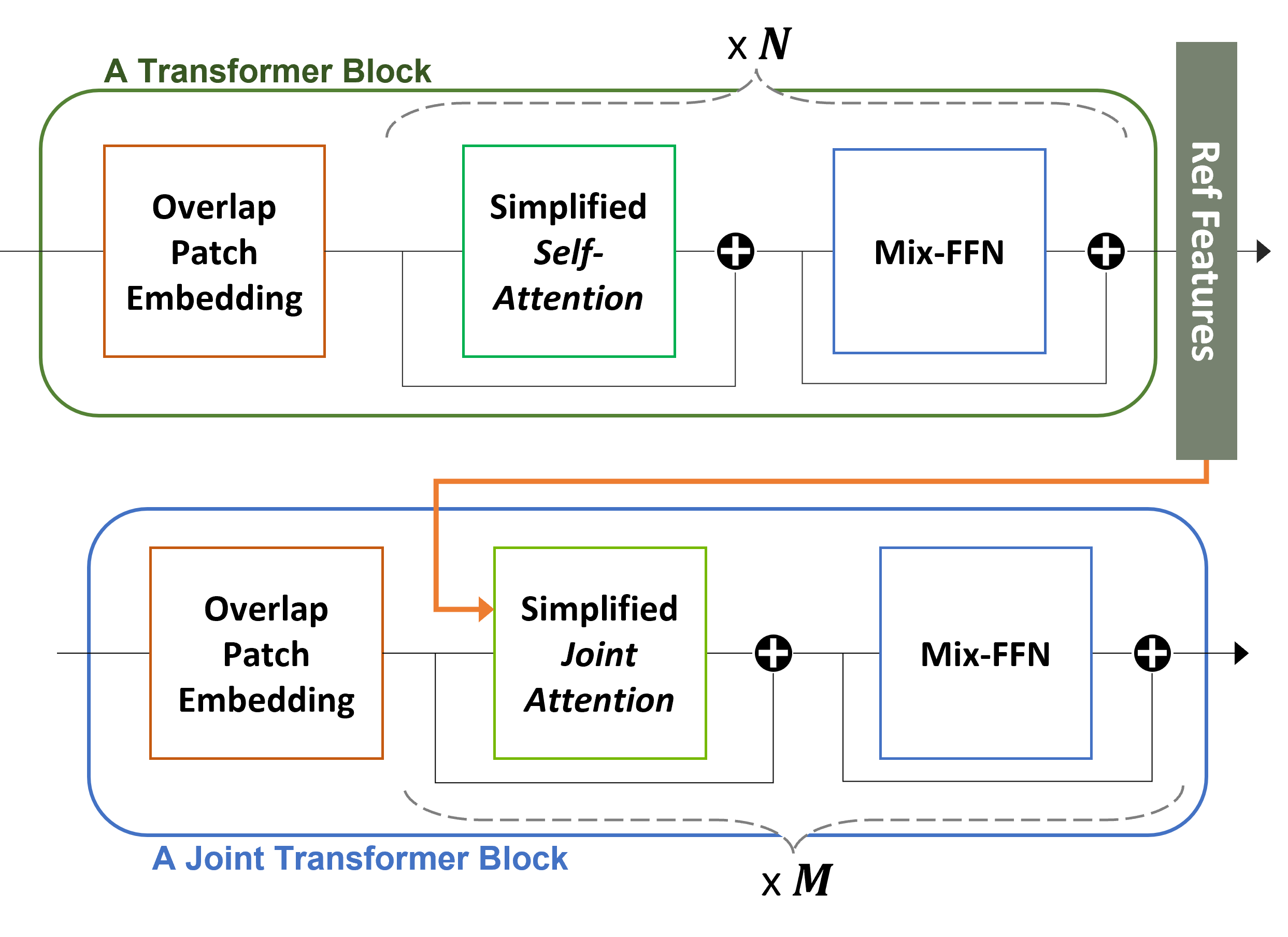}
    \vskip -0.1in
    \caption{Transformer block for Depth-Net (top) and joint transformer block for Pose-Net (bottom). In joint transformer block, the features from Depth-Net are used to computed joint attention.}
    \vskip -0.2in
    \label{fig:blocks}
\end{figure}

\subsection{Simplified Transformer}
Each transformer block contains an \textit{Overlapping Patch Embedding} layer, followed by repetitive sub-blocks of \textit{Simplified Attention} and \textit{Mix-FFN} layers, as shown in Fig.~\ref{fig:blocks}.

\noindent \textbf{Overlapping Patch Embedding.}
To preserve the continuity of local image context, similar to the approach in~\cite{xie2021segformer}, we employ an overlapping patch embedding block which simply consists of a 2D convolution followed by a batch normalization (BN). The size of input feature map is reduced by half in the beginning of every transformer block.

\begin{figure}
    \centering
    \begin{subfigure}[b]{\linewidth}
        \centering
        \includegraphics[width=\textwidth, trim={2.5mm 0 2.4mm 0}, clip]{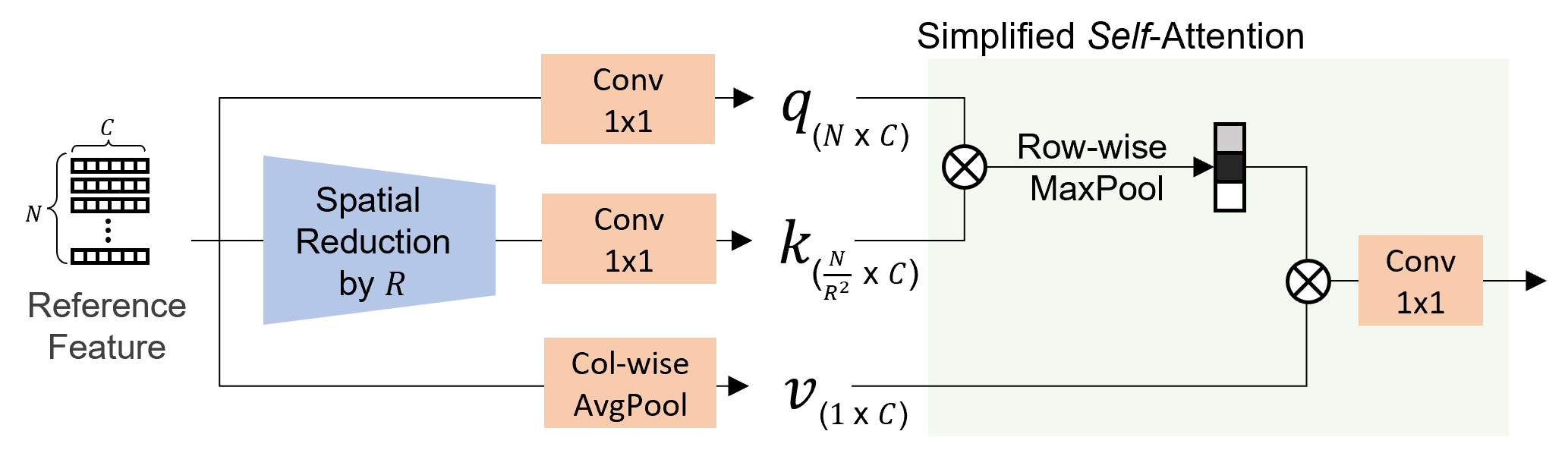}
        \vspace{-2mm}
        \caption{Simplified \textit{Self}-Attention in Depth-Net}
        \label{fig:simp_self_atten}
    \end{subfigure}
    \begin{subfigure}[b]{\linewidth}
        \centering
        \includegraphics[width=\textwidth, trim={2.5mm 0 2.4mm 0}, clip]{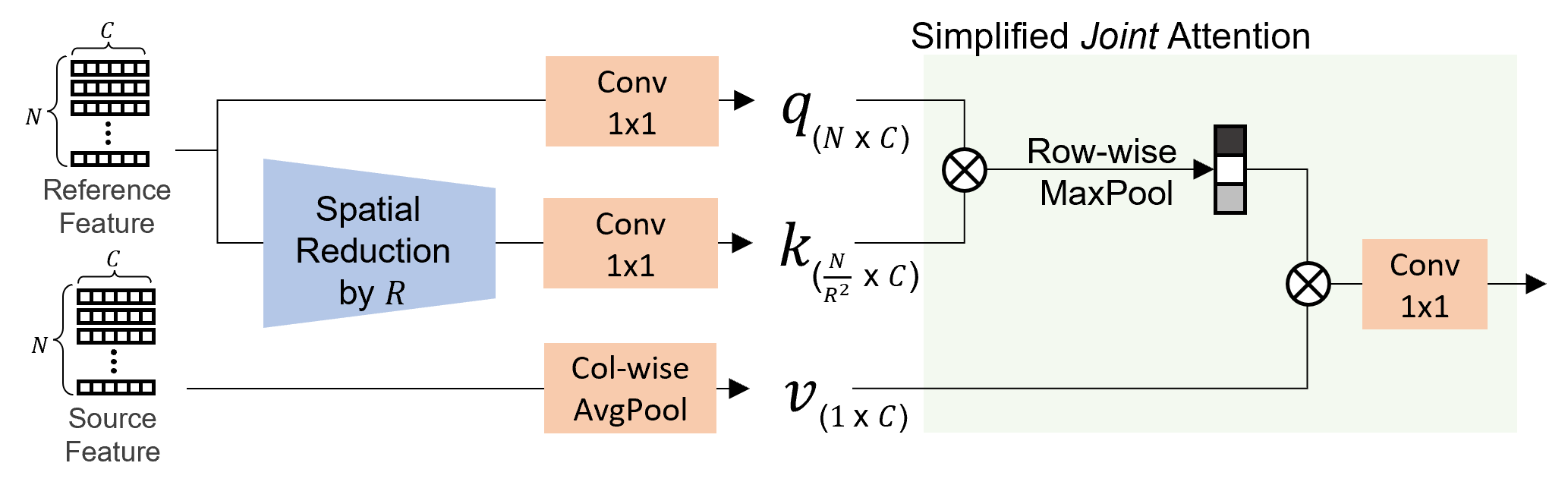}
        \vspace{-2mm}
        \caption{Simplified \textit{Joint} Attention in Pose-Net}
        \label{fig:simp_joint_atten}
     \end{subfigure}
    \vspace{-2mm}
    \caption{Simplified attention modules. The softmax used in attention is replaced by max pooling and the $v$ values are obtained by applying average pooling from input features.}
    \label{fig:simplified_attention_modules}
\end{figure}

\noindent \textbf{Simplified Attention.}
To reduce the complexity in the attention module, we made several improvements. First, we apply the sequence reduction process~\cite{pvt, xie2021segformer} with a reduction ratio $R$ such that the size of $K$ is reduced from $N \times C$ to $\frac{N}{R^2} \times C$. Secondly, instead of using a learnable layer to obtain $v$, we apply a column-wise average pooling to reduce $v$ from $N \times C$ to $1 \times C$. In addition, the softmax operation in attention yields gradients that mainly flow through maximum outputs, which is similar to the max-pooling operation. Therefore, we replace softmax with row-wise max pooling to remove the complexity in computing softmax. The self-attention is shown in Fig.~\ref{fig:simp_self_atten}, which is used in the Depth-Net. For the Pose-Net, we propose the joint attention that receives $q$ and $k$ features from corresponding blocks in Depth-Net, and $v$ from features in the Pose-Net, as shown in Fig.~\ref{fig:simp_joint_atten}. The feature sharing allows additional gradient signals for Depth-Net when it is jointly trained with Pose-Net \cite{wang2018learning_connectivity}. With the aforementioned improvements, the computation in the attention is greatly reduced as compared to other vision transformers such as that in~\cite{xie2021segformer}.

\begin{figure}
    \centering
    \includegraphics[width=0.77\columnwidth]{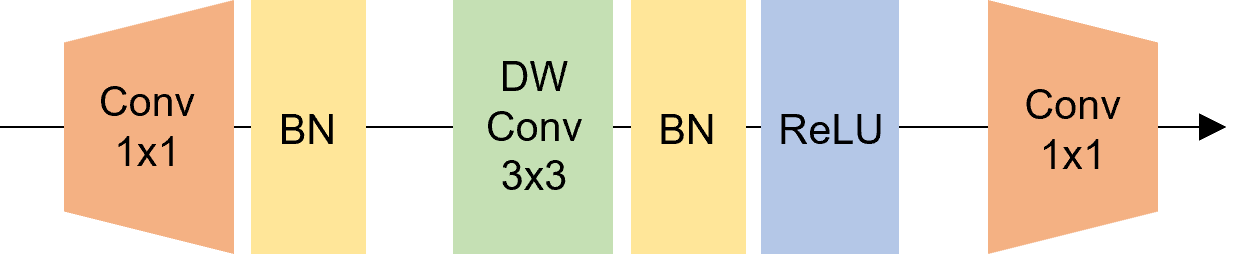}
    \vspace{-2mm}
    \caption{Mix-FFN in the transformer blocks.}
    \vspace{-4mm}
    \label{fig:mix-ffn}
\end{figure}

\noindent \textbf{Mix-FFN.} 
Instead of applying MLP after attention modules, we employ mix-FFN layer instead. The structure of our mix-FFN is shown in Fig.~\ref{fig:mix-ffn}, which is implemented differently from that in ~\cite{xie2021segformer}. The BN layers after 1x1 convolution and depth-wise convolution are crucial in stabilizing the training, and ReLU substitutes the GELU activation for reduced computation.

\noindent \textbf{Elimination of Layer-Norms.}
Layer normalization (LN) is widely adopted in transformer-based networks. However, LN requires statistics of inputs being computed on the fly, which imposes additional cost at inference time. On the other hand, BN uses the accumulated statistics from training and avoids such computation in the inference, and thus can be more favorable in transformers~\cite{Yao_2021_ICCV, shen2020powernorm}. Note that simply replacing LN with BN would cause the training to diverge, therefore our LN-free networks are implemented as the following: \textbf{1}. Two BNs are inserted in the mix-FFN layers as shown in Fig.~\ref{fig:mix-ffn}; \textbf{2}. LNs in the spatial reduction layer~\cite{pvt, xie2021segformer} are removed, and the attention layers are also free of normalization; \textbf{3}. A BN layer is placed after every joint transformer block in the Pose-Net, while the transformer blocks in the Depth-Net encoder are not followed by any normalization. Such modifications intend to lessen BN computations during inference while still stabilizing the training, since only Depth-Net is needed for inference.

\noindent \textbf{Progressive Decoder.}
Inspired by the concept of feature pyramid network (FPN)~\cite{fpn}, the decoder in the Depth-Net receives outputs from all transformer blocks in the encoder, and combines them in a progressive manner as shown in Fig.~\ref{fig:dest}. Bilinear interpolation is used to upsample the feature maps, and the size of the output depth map from the decoder is equal to the input size.


\noindent \textbf{Network Composition.}
To this end, our networks are largely composed of very simple and basic operations, namely \textit{convolution}, \textit{depth-wise convolution}, \textit{batch normalization}, \textit{ReLU}, \textit{max-pooling}, and \textit{avg-pooling}. Our design is driven in such a way since these operations have been well supported and optimized on GPU-based platforms, therefore making our networks hardware-friendly and efficient. 


\begin{table*}[!ht]
\footnotesize
    \centering
	\begin{tabular}{cccccccccc}
	\specialrule{1pt}{1pt}{1pt}
 	\hline
 	Depth-Net & Pose-Net & Connectivity & Dataset & Abs Rel$\downarrow$ & Sq Rel$\downarrow$ & RMSE$\downarrow$ & RMSE\textsubscript{log}$\downarrow$ & \#MParams$\downarrow$ & \#GMACs$\downarrow$ \\
 	\hline
 	Monodepth2~\cite{monodepth2} & ConvNet & No & K & 0.132 & 1.044 & 5.142 & 0.210 & 14.84 & 8.04\\
 	Monodepth2~\cite{monodepth2} & ConvNet & No & IN+K & 0.115  & 0.903 & 4.863 & 0.193 & 14.84 & 8.04\\
	PackNet-SfM~\cite{packnet} & ConvNet & No & K & 0.111 & 0.800 & 4.576 & 0.189 & 128.29 & 205.49 \\
	PackNet-SfM~\cite{packnet} & ConvNet & No & CS+K & 0.108 & 0.758 & 4.506 & 0.185 & 128.29 & 205.49 \\
	SETR-MLA32~\cite{zheng2021rethinking} & ConvNet & No & K & 0.184 & 1.669 & 6.407 & 0.257 & - & - \\
	SegFormer-B0~\cite{xie2021segformer} & ConvNet & No & K & 0.139 & 0.981 & 5.657 & 0.205 & 3.82 & 4.21\\
	SegFormer-B3~\cite{xie2021segformer} & DEST-B3 & No & K & 0.119 & 0.803 & 5.033 & 0.182 & 47.32 & 35.02 \\
	SegFormer-B3~\cite{xie2021segformer} & DEST-B3 & Yes & K & 0.113 & 0.798 & 4.917 & 0.179 & 47.32 & 35.02 \\
	SegFormer-B3~\cite{xie2021segformer} & DEST-B3 & Yes & CS+K & 0.105 & 0.794 & 4.707 & 0.172 & 47.32 & 35.02 \\
	\hline
    DEST-B0 & DEST-B3 & Yes & CS+K & 0.116 & 0.910 & 4.982 & 0.219 &  4.68 & 4.82 \\
    DEST-B1 & DEST-B3 & Yes & CS+K & 0.115 & 0.868 & 4.724 & 0.207 & 10.12 & 14.37 \\
    DEST-B2 & DEST-B3 & Yes & CS+K & 0.108 & 0.831 & 4.636 & 0.181 & 16.03 & 17.19 \\
    DEST-B3 & DEST-B3 & Yes & CS+K & 0.103 & 0.796 & 4.410 & 0.170 & 19.71 & 19.78 \\
    DEST-B4 & DEST-B3 & Yes & CS+K & 0.098 & 0.767 & 4.285 & 0.168 & 38.53 & 27.98 \\
    DEST-B5 & DEST-B3 & Yes & CS+K & 0.095 & 0.752 & 4.207 & 0.165 & 45.95 & 31.50 \\
	\hline
    \end{tabular}
    \vspace{-2mm}
    \caption{Quantitative results on the KITTI dataset~\cite{kitti}. Connectivity refers to the feature sharing between Depth-Net and Pose-Net. The \#MParams and \#GMACs are from Depth-Net only since Pose-Net is not needed in inference. CS+K and IN+K refer that the given model is pre-trained either with CityScapes \cite{Cordts2016Cityscapes} or ImageNet \cite{deng2009imagenet} data.}
    \vspace{-2.5mm}
   \label{kitti}
\end{table*}

\section{Experiments}

\noindent \textbf{Implementation Details.}
Our models are implemented in PyTorch 1.10 and trained with 8 NVIDIA V100 GPUs. Adam optimizer with an initial learning rate of $4 \times 10^{-5}$ is used for training. For Depth-Net, a batch size of one is used, and for Pose-Net, training sequences are generated with a stride of 2, meaning that $I_{t-1}$ and $I_{t+1}$ are concatenated with $I_{t}$ as an input. The input resolution to our models and benchmark models is $640 \times 192$. 

Similar to~\cite{xie2021segformer}, we implement six variants of DEST models. Specifically, our lightweight models DEST-B0 and B1 have a configuration of (2, 2, 2, 2), each of which indicates the number of attention and mix-FFN layers in a transformer block, as shown in Fig.~\ref{fig:blocks}. Those settings are (3, 3, 6, 3) for B2, (3, 6, 8, 3) for B3, (3, 8, 12, 5) for B4, and (3, 10, 16, 5) for the largest model B5. We set the numbers of output feature maps to (32, 64, 160, 256) for B0 and (64, 128, 250, 320) for B1-B5. For Pose-Net, we use B3 in all experiments.



\begin{figure}
    \captionsetup[subfigure]{labelformat=empty}
    \begin{subfigure}{0.327\columnwidth}
    \centering 
    \caption{Inputs}
    \vspace{-1mm}
    \includegraphics[width=\columnwidth]{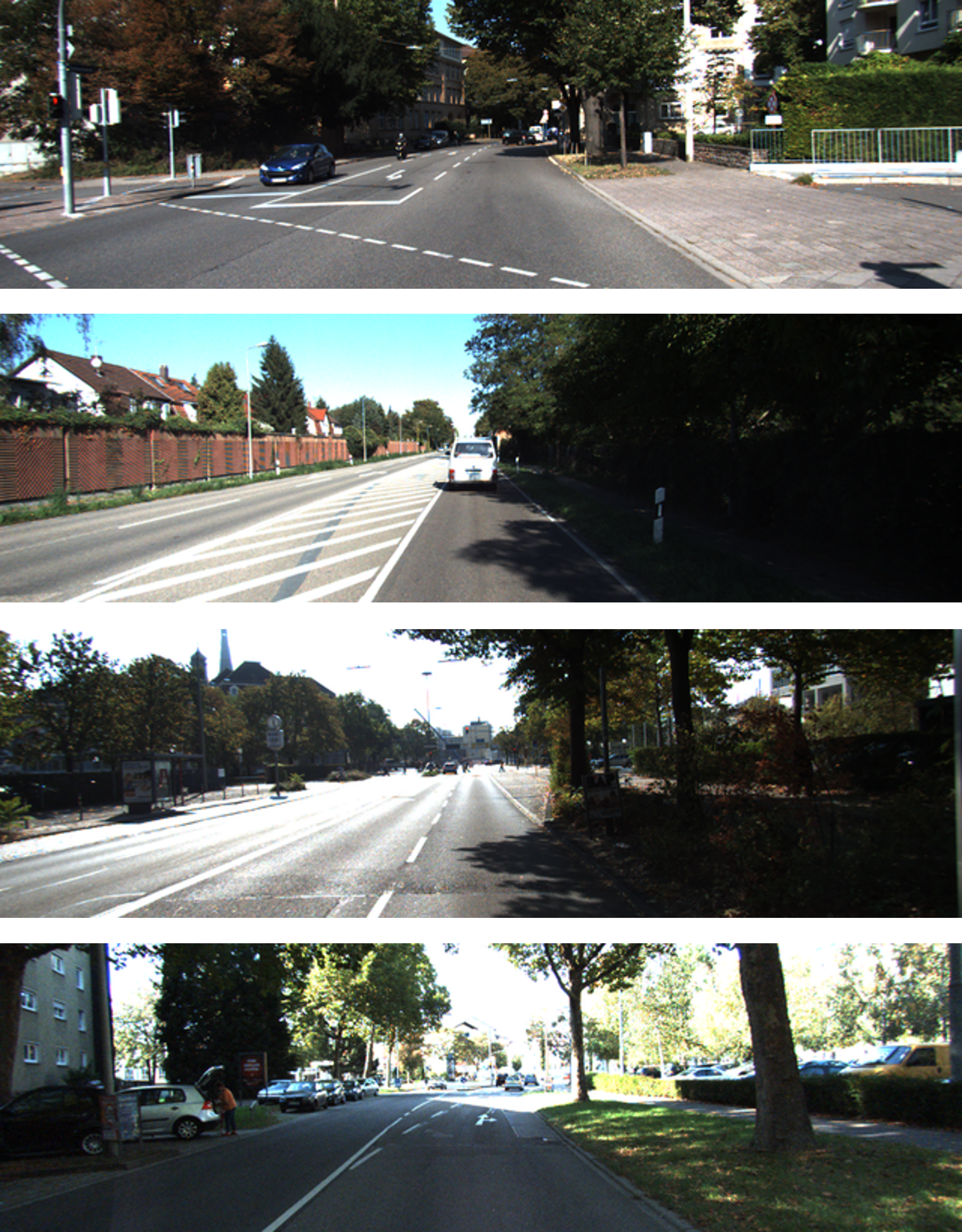}
    \end{subfigure}
    \begin{subfigure}{0.327\columnwidth}
    \centering 
    \caption{DEST-B3}
    \vspace{-1mm}
    \includegraphics[width=\columnwidth]{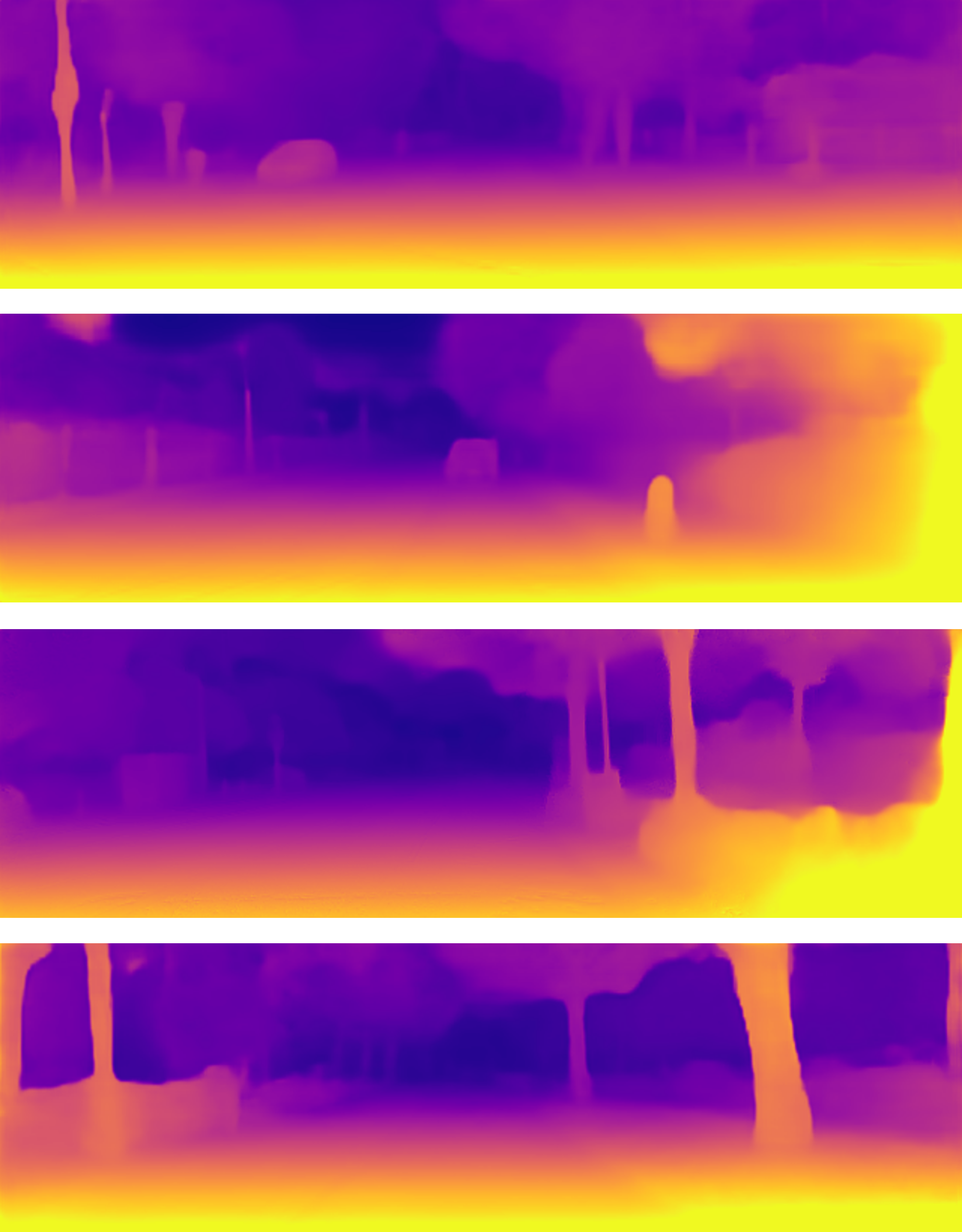}
    \end{subfigure}
    \begin{subfigure}{0.327\columnwidth}
    \centering 
    \caption{PackNet-SfM~\cite{packnet}}
    \vspace{-1mm}
    \includegraphics[width=\columnwidth]{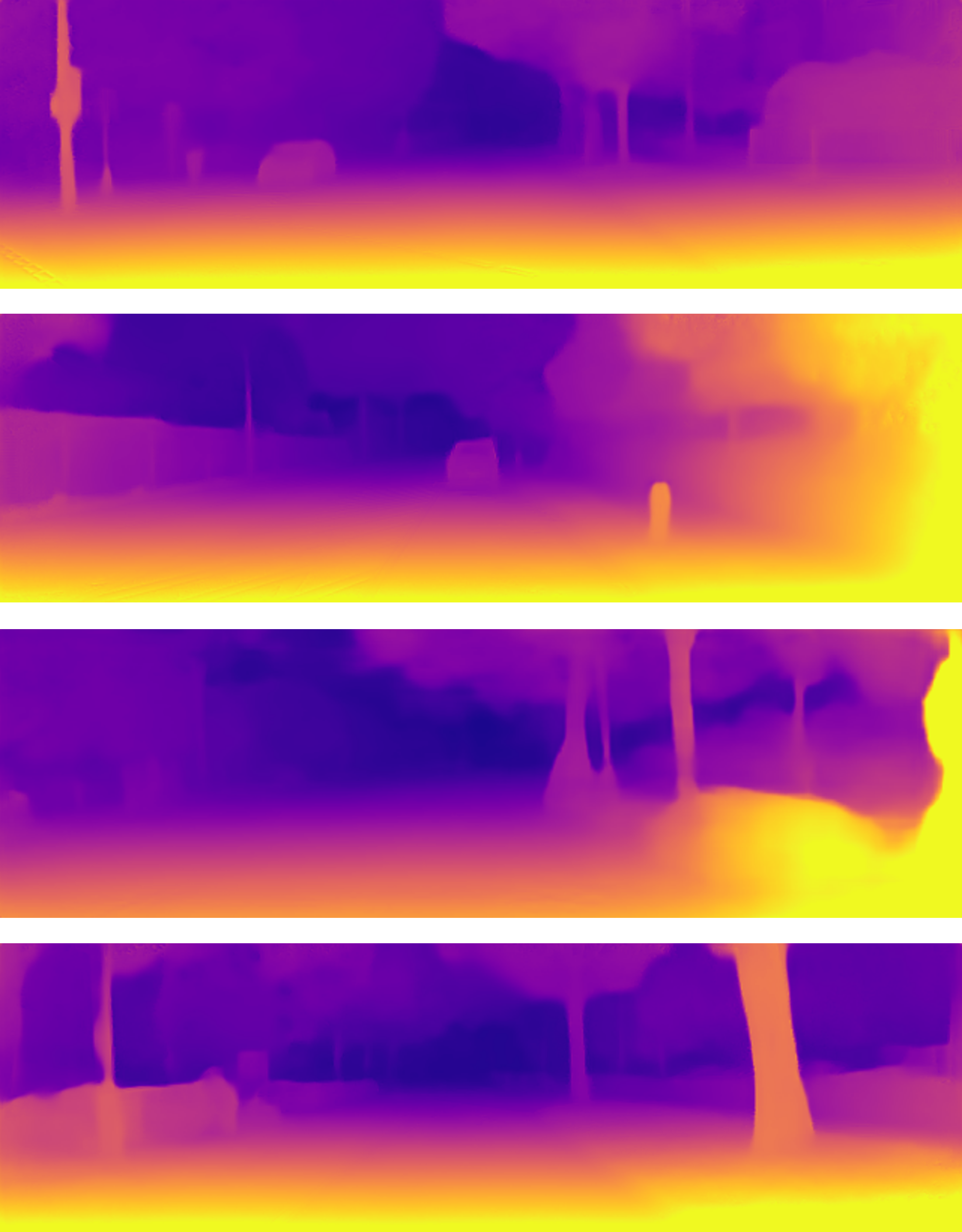}
    \end{subfigure}
    \vspace{-2mm}
    \caption{Samples of predicted depth maps.}
    \vspace{-3mm}
    \label{fig:kitti_depth}
\end{figure}

\noindent \textbf{Accuracy.}
We benchmark on KITTI dataset~\cite{kitti} and report the metrics described in~\cite{eigen2014depth}. The quantitative results are summarized in Table~\ref{kitti}. Compared to the current state-of-the-art PackNet-SfM~\cite{packnet} with pre-training, our Depth-Net B3 achieves over 85\% reduction in the number of parameters and 90\% reduction in the number of MACs, while outperforming in most metrics by a large margin. Even the smallest model B0 yields competitive results compared to other methods. The proposed connectivity between Depth-Net and Pose-Net helps improve the accuracy, as suggested by the results of baseline Depth-Net using SegFormer-B3. Some sample predicted depth maps are shown in Fig.~\ref{fig:kitti_depth}, in which fine details and structures are observed.



\noindent \textbf{Latency.}
We compare the latency of DEST-B3 to PackNet-SfM~\cite{packnet} in their original PyTorch implementations as well as the inference results using NVIDIA TensorRT 8.2 library in FP32 and FP16 precisions. All the latency numbers in our experiments were obtained on an NVIDIA V100 GPU. Table~\ref{table:latency} summarizes the runtime with a batch size of one. Compared to PackNet-SfM~\cite{packnet}, our model runs notably faster in all settings. With the help from TensorRT, the latency of our model further improved from 18.31ms to 12.36ms in FP32 precision. When FP16 is enabled, the latency of our model drops to 8.87ms. The improvements in latency resonate with the reduction in model size and computation, affirming the efficacy of our design choices.

\begin{table}
    \footnotesize
    \centering
	\begin{tabular}{cccc}
	\specialrule{1pt}{1pt}{1pt}
 	\hline
 	& PyTorch$\downarrow$ & TensorRT FP32$\downarrow$ & TensorRT FP16$\downarrow$\\
 	\hline
	PackNet-SfM\cite{packnet} & 64.76 & 42.54 & 20.54 \\
	DEST-B3 & 18.31 & 12.36 & 8.87 \\
 	\hline
    \end{tabular}
    \vspace{-2mm}
    \caption{Comparison of end-to-end inference latency in ms using PyTorch and TensorRT with different precisions.}
    \vspace{-4mm}
    \label{table:latency}
\end{table}

\noindent \textbf{Generalization to Semantic Segmentation.}
We directly apply our model to semantic segmentation with a simplified decoder that outputs a segmentation map with half size of that of the input by removing the last re-sample block. We compare ours to SegFormer~\cite{xie2021segformer} with both models trained from scratch on the CityScapes dataset~\cite{Cordts2016Cityscapes} . Table~\ref{table:semseg} shows the results with an input resolution of $1024 \times 1024$. Without bells and whistles, our model achieves better segmentation quality with greatly reduced latency.

\begin{table}[!t]
    \footnotesize
    \centering
	\begin{tabular}{ccccc}
	\specialrule{1pt}{1pt}{1pt}
 	\hline
 	& mIoU$\uparrow$ & \#MParams$\downarrow$ & \#GMACs$\downarrow$ & Latency (ms)$\downarrow$\\
	\hline
	SegFormer-B0 & 62.57 & 3.82 & 30.58 & 28.58 \\
	DEST-B0 & \textbf{63.12} & \textbf{3.58} & \textbf{22.75} & \textbf{13.94} \\
	\hline
	SegFormer-B3 & 72.30 & 47.32 & 299.07 & 127.17 \\
	DEST-B3 & \textbf{72.58} & \textbf{18.08} & \textbf{142.78} & \textbf{51.41} \\
 	\hline
    \end{tabular}
    \vspace{-2mm}
    \caption{Comparisons with SegFormer~\cite{xie2021segformer} on the CityScapes dataset~\cite{Cordts2016Cityscapes}. Latency is obtained with FP16 precision.}
    \vspace{-4mm}
    \label{table:semseg}
\end{table}

\section{Conclusions}
In this paper, we presented a design of simplified transformer for self-supervised depth estimation. The simplifications as well as the proposed joint-attention and connectivity mechanism have shown to be effective with greatly reduced model complexity and inference latency, as compared to other benchmark methods. We also showed that our model can be directly generalized to other dense image prediction task such as semantic segmentation with improved accuracy and latency. We hope our design can serve as a practical choice for real-world applications such as autonomous driving and robotics, where both high efficiency and accuracy are demanded at the same time. In the future, we would like to further evaluate our model with quantization-aware training and inference in lower precision for further improved efficiency.

{\small
\bibliographystyle{ieee_fullname}
\bibliography{egbib}
}

\end{document}